\documentclass[runningheads]{llncs}
\usepackage{graphicx}
\usepackage{tikz}
\usepackage{subcaption}
\usepackage{sidecap}
\usepackage{comment}
\usepackage{amsmath,amssymb} % define this before the line numbering.
\usepackage{color}
\usepackage{bbm}
\usepackage[accsupp]{axessibility}  % Improves PDF readability for those with disabilities.

\usepackage{hyperref}
\usepackage{algorithm2e}
\usepackage{tabularx,booktabs,multirow}
\usepackage{wrapfig}

\begin{document}
% \linenumbers
\pagestyle{headings}
\mainmatter
\def\ECCVSubNumber{37}  % Insert your submission number here

\title{Towards an Error-free Deep Occupancy Detector for Smart Camera Parking System}  % **** Enter the paper title here

\begin{comment}
\titlerunning{ECCV-22 submission ID \ECCVSubNumber} 
\authorrunning{ECCV-22 submission ID \ECCVSubNumber} 
\author{Anonymous ECCV submission}
\institute{Paper ID \ECCVSubNumber}
\end{comment}
%******************

% \begin{comment}
\titlerunning{Towards an Error-free OcpDet for SPS}
\author{Tung-Lam Duong\orcidID{0000-0002-9459-4705}\index{Duong, Lam} \and
Van-Duc Le\orcidID{0000-0002-3333-1848} \and
Tien-Cuong Bui\and
Hai-Thien To}
\authorrunning{L. Duong}
\institute{Seoul National University, Seoul, Korea\\
\email{\{dtlam26,levanduc,cuongbt91,haithienld\}@snu.ac.kr}}
% \end{comment}
%******************
\maketitle
\newcolumntype{y}{>{\hsize=.15\hsize\centering\arraybackslash}X}
\newcolumntype{x}{>{\hsize=.2\hsize\arraybackslash}X}

\begin{abstract}
Although the smart camera parking system concept has existed for decades, a few approaches have fully addressed the system's scalability and reliability. As the cornerstone of a smart parking system is the ability to detect occupancy, traditional methods use the classification backbone to predict spots from a manual labeled grid. This is time-consuming and loses the system's scalability. Additionally, most of the approaches use deep learning models, making them not error-free and not reliable at scale. Thus, we propose an end-to-end smart camera parking system where we provide an autonomous detecting occupancy by an object detector called OcpDet. Our detector also provides meaningful information from contrastive modules: training and spatial knowledge, which avert false detections during inference. We benchmark OcpDet on the existing PKLot dataset and reach competitive results compared to traditional classification solutions. We also introduce an additional SNU-SPS dataset, in which we estimate the system performance from various views and conduct system evaluation in parking assignment tasks. The result from our dataset shows that our system is promising for real-world applications.
\end{abstract}

\section{Introduction}
According to the 2018 UN media, 68\% of the world's population will move to urban areas by 2050\cite{unmedia}. This dense population in towns and cities directly leads to an increase in the number of cars and other vehicles, which raised a major concern on the parking management on their capacities and efficiency. Letting drivers wander in the city to find an appropriate parking slot in a tight city space causes significant air pollution and wastes drivers' time and energy. It also leaves empty spaces in parking lots and varies statistic measurements on parking occupancy rate, which trouble operators to exploit their facility for revenue. In addition, these factors may worsen, particularly during peak hours when the flow density is at its maximum. For real concrete evidence, a recent report by INRIX \cite{cookson2017parking} shows that on average, a typical American driver spends 17 hours a year looking for a parking space, which can go up to 107 hours when addressing a dense population city like New York. From \cite{paidi2022co2} analysis, the exceeding of CO2 emissions can rise nearly three times due to this problem. Therefore, a stable future city needs a Smart Parking System (SPS) that can serve as a link between drivers and parking operators and benefit both sides. By suggesting optimal parking places to drivers and managing their destination, a future SPS not only minimizes vehicle emissions (via decremented delays in finding the vacant parking spot \cite{al2019smart}), but also provides operators a reliable number of customers to boost revenues through e.g. dynamic pricing \cite{polycarpou2013smart}.

Regardless of potential promises, most SPS functionalities are strictly bounded by the performance of correctly determining the occupancy of a parking lot. Hence, our current parking system relies heavily on sensors as the first layer of the system \cite{al2019smart}. However, despite its high precision, this turns out to be pricey when scaling up the parking lot size for future perspective, as each sensor (magnetometer/ultrasonic sensor)\cite{polycarpou2013smart} is designed to operate solely on a single parking spot. An effective solution for this drawback is applying computer vision(CV) to occupancy detection. A single camera can cover a multiple of parking locations and eliminate
the need for a sensor per parking spot\cite{lisboa2022systematic}. Furthermore, because most of nowadays parking lots have security cameras, it reduces installation and maintenance costs and supports multiple additional tasks for better parking management, such as wrong parking placement, abnormal behavior, and theft detection \cite{varghese2019efficient} with which sensors fail to cope. 

Although computer vision is a promising approach, no datasets exist for a full CV-SPS intention. Most popular datasets PKLot\cite{de2015pklot}, and CNRPark-Ext\cite{amato2018wireless} and their solutions \cite{varghese2019efficient,amato2017deep,valipour2016parking,nyambal2017automated,amato2018wireless} are constrained to a small number of parking lots and treat each parking spot as a binary classification image. Regarding performance, there are three main drawbacks to this type of dataset and their following research. First, it limits solutions to operate in the classification scheme solely. Second, when the number of slots increases, reliable deep-learning classification solutions \cite{valipour2016parking,nyambal2017automated} require multiple forward passes and are slow to run in real-time feedback to drivers or to stack more additional tasks. Lastly, a parking operator using solutions from this dataset must reannotation every parking spot for new installation. For example, an operator is in charge of three parking lots with at least 300 parking spots in each facility. He must perform 900 annotations to use the solution, and this procedure will be conducted again when the positions of the cameras change. Therefore, an automatic parking space localization and classification solution for a scalable CV-SPS is needed to deal with the future urban population. Moreover, when we take into the functionalities of SPS in these datasets, there is no information on the parking location or surrounding traffic in this dataset as they only focus on occupancy results. It creates a big gap between existing data and the SPS's scope in the CV paradigm. 

Aware of those flaws in current CV approaches and the importance of CV in future SPS, we proposed a complete CV-SPS with a new SPS-based dataset called SNU-SPS. We provide a different solution to the first layer of SPS by treating the parking lot occupancy as an object-detection task and propose a new variant model architecture called OcpDet. Changing the scope to object-detection lifts the system's performance to real-time operation and produces results frame by frame. However, most object detector is not error-free and can be potentially wrong in a real-world inference. Hence, to maintain a reliable SPS, we provide SPS with a result filter in the second layer by addressing results from two modules of our object detector: the spatial estimator module and the training error module. While the spatial estimator module creates a predicted parking region and compares it with the model output to form the spatial error, the training error module catches the training difference of an inference frame and assigns an error. Incorrect inferences are recorded, marked as unusable information, and collected for fine-tuning and retraining the detection model. Then finally, only correct/believable detection results are stored and analyzed in the middleware layer and transparent to operators and drivers. From this layer, we can support optimal routing for drivers and alert operators about upcoming occupied parking slots at the top layer. SNU-SPS dataset is created to support this idea of the system. It contains parking slots captured from multiple parking lots at various angles, ranges, and positions with different light and contrast settings to train the detectors. Furthermore, we provide parking lot GPS with its occupancy rate and surrounding traffic information for system performance analysis.

We extensively test our proposed system on our dataset for efficiency evaluation and conduct detection measurements with the popular parking datasets PKLot for a detailed benchmark. The results from our experiment raise a competitive performance compared to exhaustive classification methods.
\section{Related Work}
\subsection{Automatic Parking Occupancy Detection}
Because most previous work focuses on solving the SPS occupancy task as image classification from datasets \cite{de2015pklot,amato2018wireless} with manual label mask/grids location of a parking lot, none of available datasets could be found for automatic parking space detection. It leads to a small amount of effort on this topic. To the best of our knowledge, there are currently two main approaches to this topic: a mask-based method and a detector-based method.

A mask-based method aims to provide parking patches directly from captures and perform binary classification for the occupancy. The perspective transformation method \cite{bohush2019extraction,bura2018edge,nieto2018automatic} is usually used in this scheme to bring the parking lot to a 2D grid presentation. Therefore, it can save time for self-annotating parking locations and exploit the classification machine-learning and deep-learning backbones. However, since the perspective projection process is highly dependent on the camera setting to the parking lot, classification models need to be retrained for different camera settings, which questions the scalability of those methods. Notice this behavior, \cite{li2017uav} has introduced a GAN approach that generates the parking place's masks from a team of drones, but there is no comprehensive measurement of the correctness of these masks. In addition, this method requires a top-view capture of the lot, making it unrealistic for indoor parking facilities.

In contrast to mask-based solutions, detector-based approaches perform detection and classification tasks in a single process by a CNN architecture instead of separating them into two processes, which maintains the flexibility and fast inference for CV-SPS infrastructure. The CNN architecture in this realm regresses a parking slot as a foreground or a region of interest and optimizes its classification score. This procedure can be classified into two-stage and one-stage detectors. While the two-stage detector, such as Faster RCNN \cite{ren2015faster} focuses on the first stage to propose the regions of interest and performs classification on those regions in the second stage, the one-stage detector combines both tasks by grid-anchor regressions. However, because an empty or small parking slot is easily confused as a part of the image's background, both of these architectures face a lot of flaws \cite{padmasiri2020automated}. Most recent works \cite{kirtibhai2020faster} only used detectors to find a parked car in the parking lot and determine the occupancy rate by a preowned parking lot's capacity and location. This approach relaxes the problems into the well-known car detection, but it limits the extension of the SPS for letting drivers know the location of the parking slot. Recent developments in new architectures such as YOLO \cite{bochkovskiy2020yolov4}, and RetinaNet \cite{lin2017focal} have opened some flexibilities in the small object capture. The idea of using a drone's captures is also used by \cite{hsieh2017drone}. The author performs the car detection at top-views by Faster RCNN and YOLO and combines it with the layout proposal. This method faces the same drawback as \cite{li2017uav} and is restricted for car detections. For a complete occupancy detection from a detector, only \cite{padmasiri2020automated} has conducted on a RetinaNet on PKLot \cite{de2015pklot} dataset. However, the results show much confusion between moving cars and occupied parking slots. While the main reason for this inefficiency is the nature of the PKLot dataset itself (partial area of the parking lot is annotated), the method's performance can be improved if there is an attention mechanism on the parking lot region. In addition, there is a potential non-optimized model design as there is no information on the grid-anchor feature selection. 
\subsection{Deep Model Uncertainty}
As a model performance is a reflection of the coverage of the training dataset, recent research tends to capture model performance in the wild by measuring its uncertainty/stability or contradiction. While \cite{kao2018localization,aghdam2019active} aimed to create this uncertainty from the outputs by comparing an image and its noise version, \cite{feng2019deep,haussmann2020scalable} measured the stability of the prediction of the same image from different models' inferences. However, these methods are not designed for online inference and require some delay for a model judgment. Only \cite{yoo2019learning} can achieve the model uncertainty within a single forward pass inference. Instead of replicating the contradiction from inferences, the author trained the model with a different head to predict this uncertainty directly. Even though it sounds naive and simple, his approach has shown promising results on general object detection datasets.
\begin{figure}[!htbp]
\centering
\includegraphics[width=\textwidth]{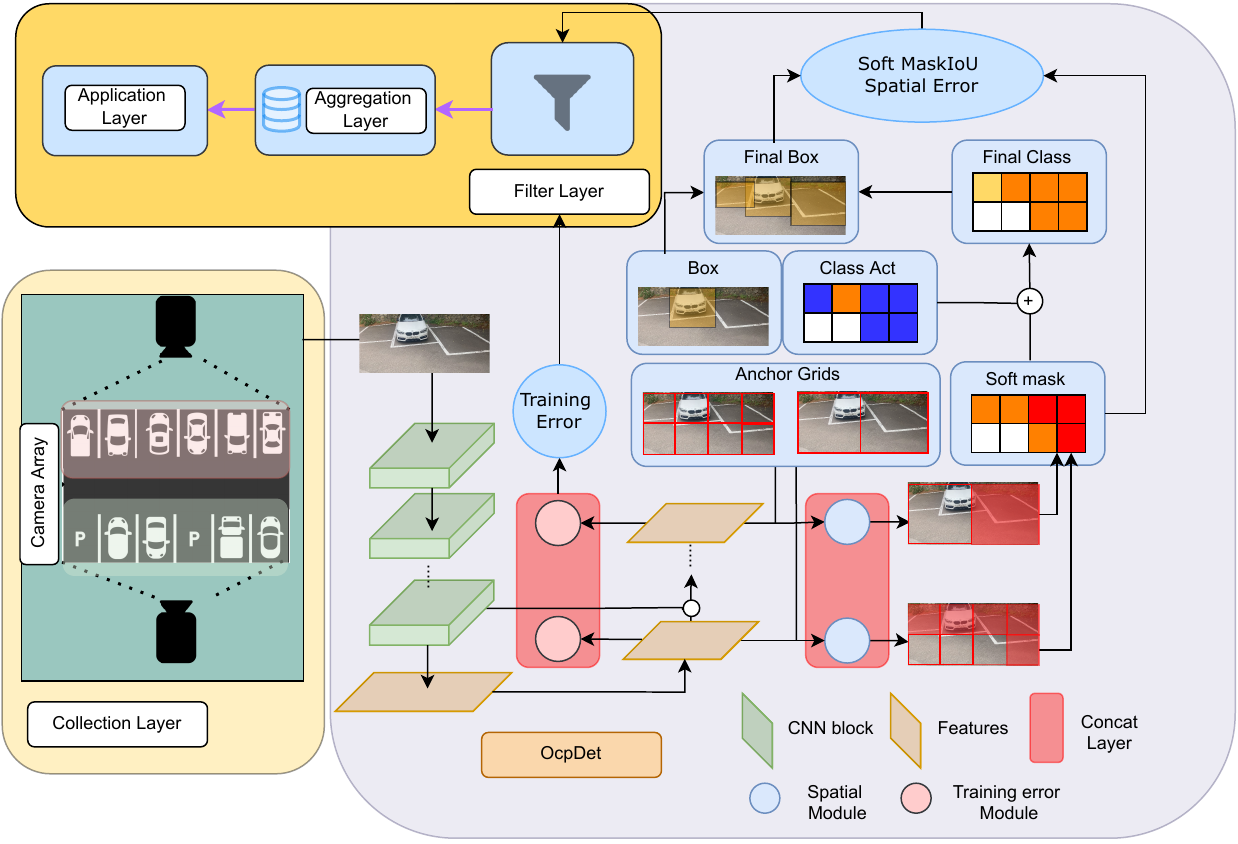}
\caption{Our CV-SPS Overall Architecture}
\label{fig: sps_architecture}
\end{figure}
% \vspace*{-15pt}

\subsection{Qualitative Comparison Between Existing Approaches and the Proposed System}
Aware of the lack of discovery in the current approaches for the CV-SPS system and the need for an efficient detector-based method, we provide a compact detector-based solution with its corresponding dataset. To our knowledge, our CV-SPS is the first end-to-end solution for automatic parking occupancy detection. Our detector-based solution inherits the feature pyramid structure from RetinaNet \cite{lin2017focal} with two additional module heads: the spatial module and the training error module. While we inherit \cite{yoo2019learning} as the training error module for training error information, we design the spatial estimator module as a parking region proposal to compare the spatial error. It can be understood as a foreground filter, aiming to filter out detection not covered in the predicted foreground region. Hence, this module provides the model with better localization attention, which \cite{padmasiri2020automated} did not address.

\section{Proposed Method and Dataset}

\subsection{Overall System Architecture}
As shown in Fig \ref{fig: sps_architecture}, we divide a parking lot into sectors to create a scalable and efficient CV-SPS. Each sector is controlled and well-observed by a camera and has non-overlapped observing areas among cameras. This constraint reduces the complexity of the problem by duplicating observation or high occlusion. Assuming the parking lot can be set up with this requirement, our overall system architecture consists of 4 layers: a collection layer, a filter layer, an aggregation layer, and an application layer. The collection layer is responsible for gathering the detection results from distributed cameras in the parking lot as well as their potential error info during inference. Then, these results are propagated to the filter layer to cleanse for reliable results. Non-trusted results are masked out as non-usable spaces. This filtered information is stored in the aggregation layer that acts as middleware of the system. From this layer, the application layer can receive reliable SPS support. Users can have a transparent measurement of the current occupancy capacity of a parking lot, while model engineers can access and inspect poor performance behavior in specific sectors. Especially, optimal routings and parking assignments can execute with high precision by correctly capturing vacant spots.

\subsection{Dataset}
\begin{figure}[!htbp]
\centering
\includegraphics[width=\textwidth,trim={0 2.5cm 0 0},clip]{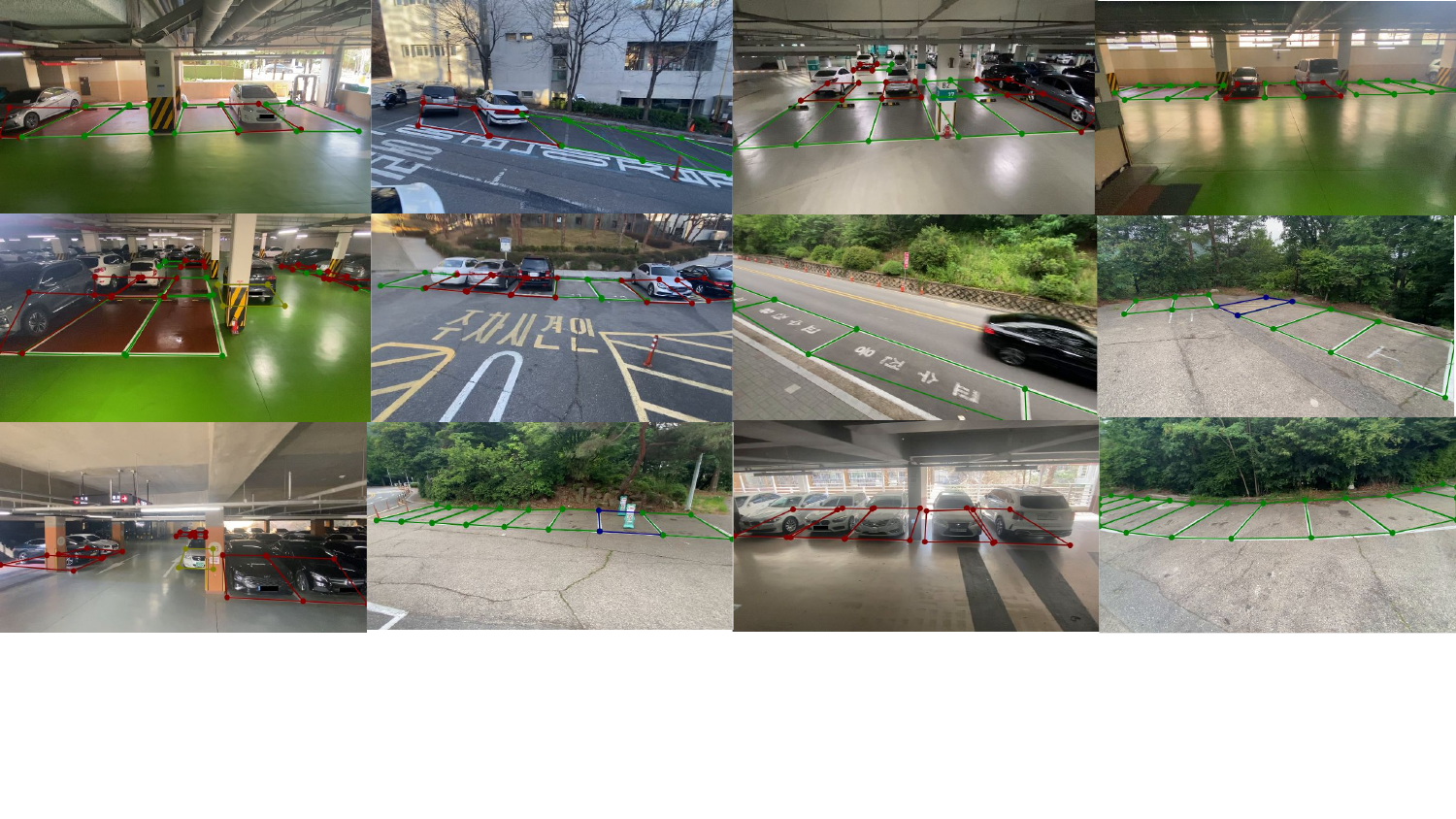}
\caption{SNU-SPS dataset images representation from various indoor and outdoor views. Annotation colors (Red: Occupied, Green: Available, Blue: Restricted and Yellow: Illegal)}
\label{fig: dataset}
\end{figure}

We introduced our SNU-SPS dataset containing nearly 3500 images to support our system. Those images are captured from various views, heights (1-3m), and light conditions in indoor and outdoor parking lots. Each parking lot has different parking spot background colors. The total images were manually checked, labeled, and attached by GPS to the corresponding parking slot. The protocol used to construct the SNU-SPS dataset is composed as follows:

\textbf{Image Acquisition:} All images are captured with a full-HD resolution. For the training set, it is captured randomly for one month in 15 parking lots. Meanwhile, the test set is captured consecutively in 6 parking lots from 3-6pm through 5 working days. It should be noted that none of the 6 parking lots are in the training set. Moreover, test samples contains various weather conditions (sun/rain/cloudy) and has corresponding surrounding traffic measurements from the open government website. 

\textbf{Labeling:} For each parking sector, images were labeled as \textit{available/ occupied/ illegal/ restricted} of each parking space. Each annotation is covered by four keypoints that specify for the localization of a parking lot. We formulate the wrapping bounding boxes for the detector from these key points. Especially, we provide optional image masks for the test set to filter out overlapping areas and non-important localization among capture among parking lots. The intention is to maintain the system's constraints and preserve a better parking assignment benchmark.
\begin{table*}[htbp]
    \centering
        \begin{tabularx}{\textwidth}{c *{8}{y}}
            \toprule
            \multirow{2}{*}{Set Type} & Total  
            & Total & \multicolumn{4}{c}{Classes} \\ 
            \cmidrule(l){4-7}  & Images & Labels & Available & Occupied & Illegal & Restricted\\
            \midrule
            Train & 2848 & 18263 & 7229 & 10596 & 396 & 42\\
            Test & 574 & 2747 & 1291 & 1336 & 36 & 84\\ 
            \bottomrule
        \end{tabularx}
    \caption{Training and Testing Sets}
    \label{tab:my_label}
\end{table*}
\subsection{Automatic Parking Occupancy Detector}
When addressing parking occupancy as an object detector, the most arousing problem is the confusion of the parking slots with the image background information such as moving cars or blank spots. The only meaningful visual information is the thin lines separating spots. However, it is usually missed at the lower level of deep neural networks. 

\textbf{RetinaNet:} is a promised solution due to its feature pyramid network (FPN). In short, the FPN backbone combines standard convolutional network lower features with lateral connections of early-level features. Hence, the network can construct rich, multi-scale object features, which maintain the impact of the line features in the network. However, from \cite{padmasiri2020automated} results, despite capturing good center localization, traditional RetinaNet could not expand the parking space tightly when the mAP dropped dramatically from 63.64 to 4.75 when raising from 0.5 to 0.75 precision. This might be non-optimized anchor grid features and lack of a location attention mechanism. Moreover, the Resnet backbone is quite heavy for computation and may not be able to scale up with other additional SPS tasks, limiting the scope of CV-SPS.
\begin{equation}
    L_{size} = \sum_{i=0}^{N} D^{i}_{p,k}/D^{i}_{p,c}
\end{equation}

\textbf{OcpDet:} Noticing this limitation, we replace the heavy Resnet backbone with a lightweight Mobilenet for faster inference and build up our model from this called OcpDet. We intend to give OcpDet more reference points rather than just centers and sizes of parking slot boxes. Therefore, instead of solely detecting the bounding boxes by their centers and sizes, our detector also predicts the key points of parking slots in the localization head. From this scope, we design a new loss function $L_{size}$ which aims to maximize the predicted coverage boxes to their corresponding predicted key points. We treat these $N$ keypoints as anchors which pull the box corners closer to them, where $N$ is less than 4. We denoted the distance between a box corner $p$ to the its corresponding keypoint $k$ as $D_{p,k}$ and the distance between a box corner $p$ to the center $c$ as $D_{p,c}$. When we set the keypoints as our dataset keypoints, the loss implicitly guides the model to focus more on the border of a parking slot.

\textbf{Spatial Estimator Module:} Increasing solely reference points is not enough to make our OcpDet robust with background and foreground conflict in a parking lot. Cars vs. parked places or empty spots vs. walls can be highly confused. Because our model is a single-stage object detector, it divides an image into anchor patches and represents them at different scale levels for classification prediction and localization regression. As the parking lot layout is usually aligned, equally separated, and non-overlapped among spots, it is very convenient to ideally wrap a parking slot in one single anchor patch representation. Therefore, we create a soft head estimator for the active anchors that can be considered as parking slots from each level feature generator of FPN. 

\begin{figure}[hbt]
\centering
\includegraphics[width=0.85\textwidth]{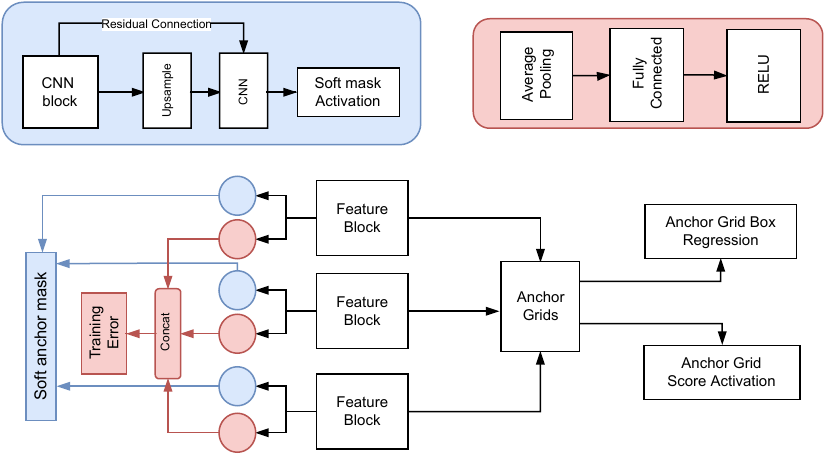}
\caption{Module Architectures for Spatial Estimatior Module and Training Error Module}
\label{fig: modules}
\end{figure}
We attach a residual convolution block for each of the $N$ feature levels and average the last channel to get a 2D map of anchor patches. Noted that, we avoid flattening the map to avoid insufficient computations as the fully connected layers can go up to huge connections for early-level features. For example, a 256x256 anchor can cost ~65k connection and additional 65 parameters for a dense layer. With that intuition, as parking slots are in the foreground, we trigger this 2D prediction by a sigmoid activation and treat it as a binary classification (Fig \ref{fig: modules}). From this activation map, we can create a soft mask for parking slot locations as a reference for the model spatial outputs. To train this activation map, we compare each map $M$ from $N$ feature levels with its corresponding classification target map $C$ in the classification head by a $\epsilon$ difference. This loss can act as a regularizer for the model to provide attention to the foreground region of the model. Moreover, as the soft mask head is not directly connected to the localization head prediction, it gives the model another degree of freedom to operate while implicitly improving the localization through top-level features.
\begin{figure}[bth]
\includegraphics[width=\textwidth,trim={0 0.5cm 1cm 0},clip]{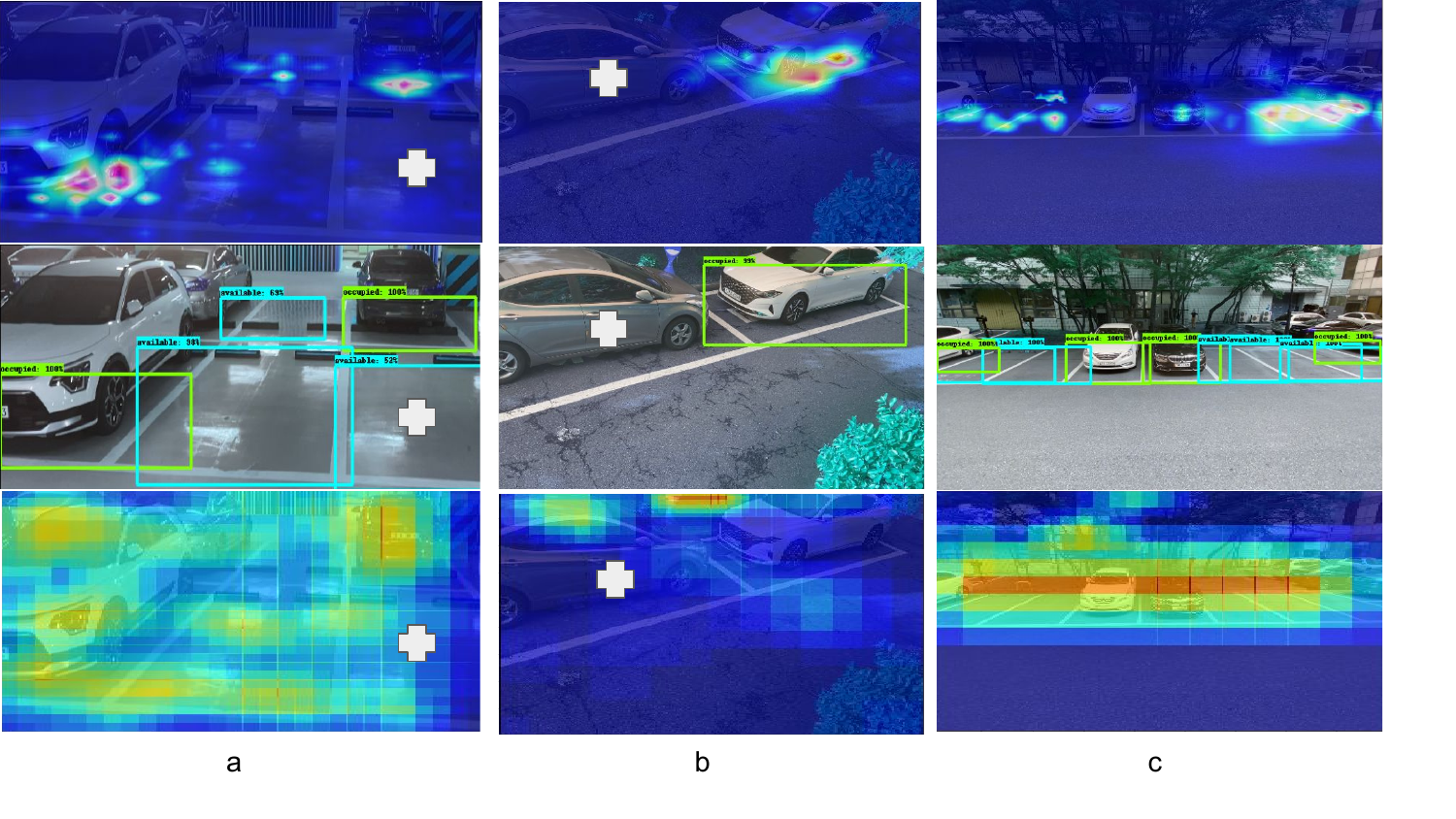}
\caption{HiRes-GradCam activations following by Image Detections (green: "occupied", blue: "available") and their coresponding soft mask heatmaps. White arrows stand for weak detection area: (a) low confidence score, (b) zero detection. c) is an exmaple of our model inference}
\label{fig: grad}
\end{figure}
To deeply understand the role of the spatial module, we investigate the way of predicting the foreground from multi-level anchor-grid features through the FPN. Regardless of the grid's anchor box size, it will be considered a foreground if the classification activation is switched on. Therefore, we look into the pixel activation map through the class activation map (HiRes-GradCam) \cite{draelos2020use} to understand which pixel contributes to the final decision. As demonstrated in Fig \ref{fig: grad}, there is a high density of pixel activations at each middle of parking slots whenever the model predicts the location class, which is reasonable as a class determination belongs to what having inside a parking border. The class confidence is reduced when the number of surrounding activation maps is dimmed Fig \ref{fig: grad}-a and disappears in Fig \ref{fig: grad}-b. This means the activation map only produces a reliable decision when it falls inside a parking spot. We notice that the spatial module's heatmap can evade this problem by adding the activation of the soft mask prediction. Following this idea, we injected the predicted classification parking-grid information into the classification head by adding its prediction to the ground confidence. We \underline{avoid multiplication} for overfitting results with the predicted soft-mask.

\textbf{Training Error Module:} as mentioned above, we use \cite{yoo2019learning} method as a model error predictor from training knowledge. In short, \cite{yoo2019learning} represents this error as predicted loss during training. The method selects pairs of training batch samples in each iteration to represent the model inference stability during training. A good pair of samples should have a low loss during training, while a poor inference will raise the loss. Hence, during training, a predictor module will try to estimate this loss by accessing multi-level features, allowing it to choose necessary information between layers to capture the loss behavior correctly. In other words, this method can also act like an abstract regularizer for the model to avoid conflict between batches. Compare to the Spatial Estimator Module in Fig \ref{fig: modules}, it shares the same feature map, but average out each feature by an average pooling before aggregation for loss prediction. Therefore, this module can operate independently from the model output performance.
\subsection{Result Filter}
Therefore, after obtaining the detection results with two additional predictions from the spatial and training error modules, the foreground $N$ detections of each sector $D$ are compared with its soft spatial activation map $S$. We compute the overlap value for each detection and select the strong-belief region by a threshold $\gamma$. From an array of overlapping ratios, we can estimate the spatial error $\textit{Err}_{spatial}$ for each frame capture detection as demonstrated in Eq. \ref{eq: Filter}. Noted that, if $\mathbbm{1}({R})$ from a detection reach near the zero bottoms of the overlap ratio, the spatial module will get rid of this detection. We combine this spatial error with the training error $\textit{Err}_{training}$ from the training error module by averaging aggregation as a final error score $\textit{Err}_{total}$ for the inference. From this score, we can eliminate whether to believe these detections results or not by thresholding our belief.
\begin{gather}
    \label{eq: Filter}
    \textit{Err}_{spatial} = 1-\frac{1}{N}\sum_{i=0}^{N}\mathbbm{1}(\frac{D_i\bigcap{S}}{D_i})\quad
    \text{s.t}\quad \mathbbm{1}({R}) = \begin{cases}
      1, & \text{if}\ R>\gamma \\
      R, & \text{otherwise}
    \end{cases}\\
    \textit{Err}_{total} = \alpha\textit{Err}_{training} + (1-\alpha)\textit{Err}_{spatial}
\end{gather}

\section{Experiment}
In this section, we demonstrate the benefits of our approach and its efficiency in a real-world scenario. We first study the impact of using the spatial module and reference keypoints for improving parking localization and compare it with the existing solutions. Second, we demonstrate the effectiveness of filtering out unstable inference to the model's actual performance. 
% Then, we look into the speed performance of the approach on an edge device level, which makes it a cheaper installment than a server set up in the production. 
Finally, we measure the impact of OcpDet's performance with the optimal routing and parking assignment task.

\subsection{Experimental Setup}
\textbf{Training settings:} For our experiments, with the scope of an efficient and quick response in the first system layer, we only addressed the single-stage object detector: SSD-Mobilenet (denoted MBN), Mobilenet-FPN (our model backbone, denoted MBN-FPN), and our OcpDet. The training engine for these models is Tensorflow Object Detection API.
% which we can serve our models on edge devices (Google Coral Dev Board and Jetson Nano) for later measurement. 
We trained OcpDet for both datasets in 25000 iterations with batchsize 48 by SGD optimizer. To keep the detection robust to the small parking spots, we selected with high-resolution 896-pixel input instead of the traditional 300-pixel or 640-pixel.

\noindent\textbf{Benchmark settings:} We used PKLot dataset \cite{de2015pklot} and our SNU-SPS dataset for the solutions benchmark. We did not use CNRPark-Ext because this dataset's parking spot border lines are faded and not consistently visible. PKLot dataset contains three sub-datasets capturing from high view: PUCPR, UFPR05, and UFPR04, which leads to a small scale of parking spots. All of these sub-dataset parking locations are partially annotated. Thus, we provide masks for each sub-dataset to clip out non-annotated parking regions to avoid false positives during training and make it suitable for the detection benchmark. We split the dataset from each sub-dataset in half for training and testing, as the PKLot authors suggested. For our SNU-SPS, we have a separate test set. However, due to a small label of \textit{illegal} and \textit{restricted} classes in our dataset, our test set is only addressed with two classes: \textit{occupied} and \textit{available}. In addition, because the scope of our dataset is to detect correctly in a sector, we only address medium and large ground truths in the test set. During the assignment application test, the detection will be filtered out overlapping areas by our provided masks.
\subsection{Detection Performance}
\begin{table}[h]
    % \begin{minipage}{0.6\textwidth}
        \begin{tabularx}{\textwidth}{xc*{7}{y}}
            \toprule
            \multirow{2}{*}{Method} & \multirow{2}{*}{Test Set } & Recall & \multicolumn{3}{c}{mAP} & \multicolumn{2}{c}{Classification Score} \\ 
            \cmidrule(l){4-8} & & (0.5:0.95) & 0.5 & 0.75 & (0.5:0.95) & Occupied & Available\\
            \midrule
            \textit{OcpDet} & \multirow{4}{*}{PUCPR} & \textcolor{red}{0.88} & \textcolor{red}{0.98} & \textcolor{red}{0.98} & \textcolor{red}{0.84} & \textcolor{red}{0.98} & \textcolor{red}{0.98}\\
            MBN-FPN & & 0.76 & 0.86 & 0.85 & 0.72 & 0.83 & 0.90\\ 
            MBN\cite{sandler2018mobilenetv2} & & 0.41 & 0.48 & 0.35 & 0.31 & 0.49 & 0.46 \\
            Classifier\cite{valipour2016parking} & & - & - & - & - & \textcolor{red}{0.99} & \textcolor{red}{0.99}\\ 
            \midrule
            \textit{OcpDet} & \multirow{4}{*}{UFPR05} & \textcolor{red}{0.98} & \textcolor{red}{0.99} & \textcolor{red}{0.99} & \textcolor{red}{0.97} & \textcolor{red}{0.99} & \textcolor{red}{0.99}\\
            MBN-FPN & & 0.84 & 0.93 & 0.90 & 0.82 & 0.93 & 0.94\\ 
            MBN\cite{sandler2018mobilenetv2} & & 0.42 & 0.51 & 0.42 & 0.37 & 0.50 & 0.53\\
            Classifier\cite{valipour2016parking} & & - & - & - & - & \textcolor{red}{0.99} & \textcolor{red}{0.99}\\ 
            \midrule
            \textit{OcpDet} & \multirow{4}{*}{UFPR04} & \textcolor{red}{0.96} & \textcolor{red}{0.99} & \textcolor{red}{0.99} & \textcolor{red}{0.93} & \textcolor{red}{0.99} & \textcolor{red}{0.99}\\
            MBN-FPN & & 0.83 & 0.95 & 0.90 & 0.79 & 0.95 & 0.96\\ 
            MBN\cite{sandler2018mobilenetv2} & & 0.43 & 0.52 & 0.44 & 0.36 & 0.51 & 0.53\\
            Classifier\cite{valipour2016parking} & & - & - & - & - & \textcolor{red}{0.99} & \textcolor{red}{0.99}\\ 
            \midrule
            \textit{OcpDet} & \multirow{7}{*}{SNU-SPS} & 0.56 & 0.81 & 0.48 & 0.47 & 0.83 & 0.80\\
            $OcpDet_{spatial}$ & & \textcolor{red}{0.56} & \textcolor{red}{0.83} & \textcolor{red}{0.50} & \textcolor{red}{0.47} & \textcolor{red}{0.85} & \textcolor{red}{0.82}\\
            $OcpDet_{both}$ & & 0.56 & 0.82 & 0.49 & 0.47 & 0.84 & 0.81\\
            $OcpDet_{ll}$\cite{yoo2019learning} & & 0.55 & 0.82 & 0.49 & 0.47 & 0.83 & 0.81\\
            MBN-FPN & & 0.54 & 0.77 & 0.46 & 0.45 & 0.80 & 0.74\\
            MBN\cite{sandler2018mobilenetv2} & & 0.51 & 0.71 & 0.48 & 0.44 & 0.73 & 0.69\\
            \bottomrule
        \end{tabularx}
    % \end{minipage}
    \caption{PKLot and SNU-SPS Detection Benchmark}
    \label{tab: performance}
\end{table}
In this part, we conduct our experiment on the efficiency of our spatial module and the localization improvement of additional reference keypoints. Our main metric for the detection evaluation is mAP(mean average of precision) and recall ranging from 0.5 to 0.95 IoU(intersection over union). Meanwhile, as the classification task is the side task to benchmark with a classification approach, we address the mAP(0.5) from each class for the comparison. As results are summarized in Table \ref{tab: performance}, the performance of the localization attention leads in all data settings. We did not compare with \cite{padmasiri2020automated} solution because of his insufficient model's performance. 

\begin{figure}
\centering
\includegraphics[width=0.95\textwidth]{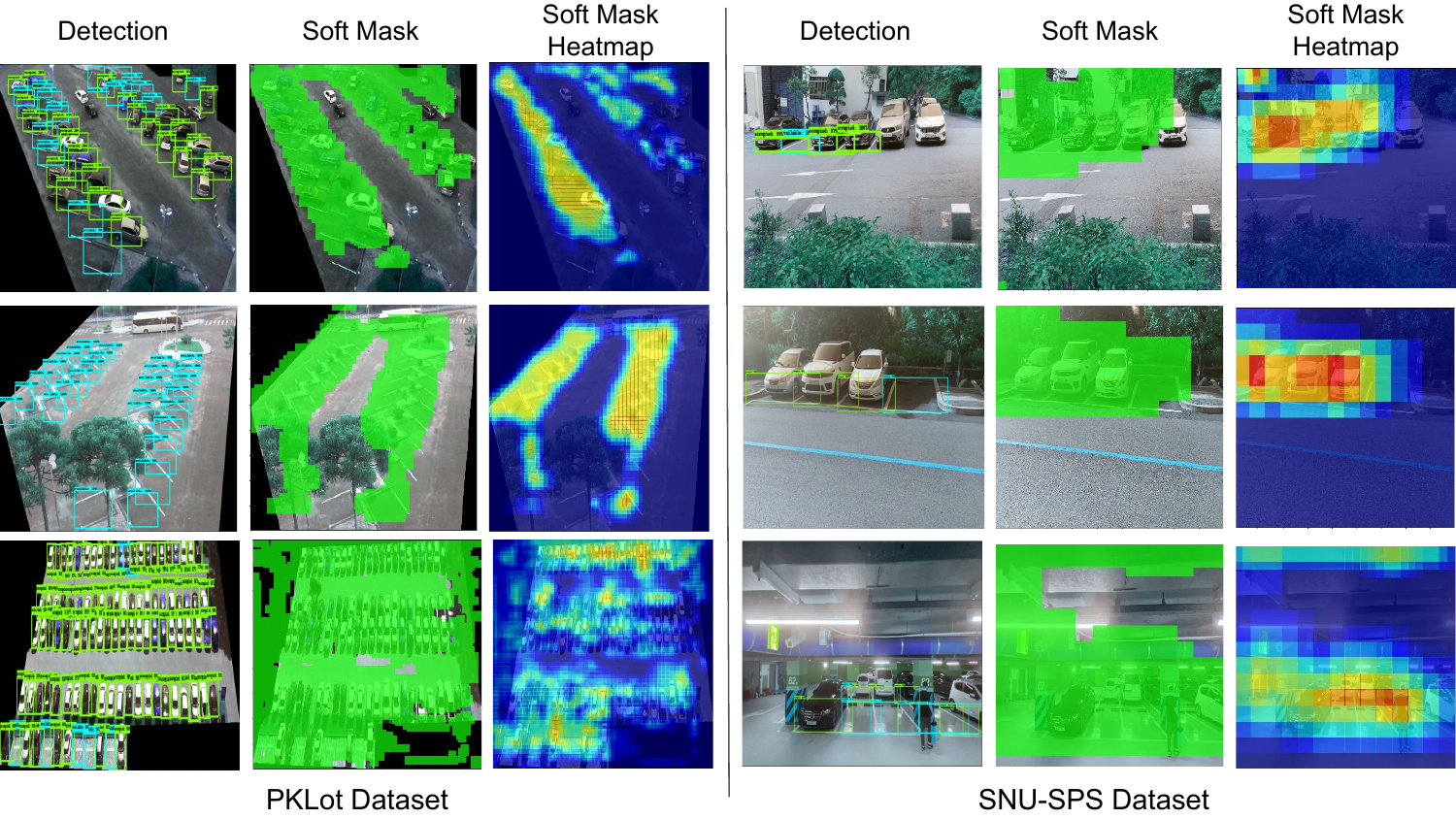}
\caption{Visualization occupancy detections, soft mask predictions and soft mask predictions heatmap on PKLot dataset and SNU-SPS dataset}
\label{fig: loc_heatmap}
\end{figure}

In the PKLot dataset, MBN struggles to learn the features because it is lack of top feature generation from the grid and lines to capture the small objects. Meanwhile, thank to the FPN, both MBN-FPN and our model outperform MBN in this dataset. Due to the fixed location of the parking lot's captures, OcpDet can strongly overfit the position of each parking space and turn into a grid classifier. As demonstrated in Fig \ref{fig: loc_heatmap}, the localization guidance from the soft mask generations helps the anchor patches avoid negative samples that do not belong to the parking area. Our model can boost the performance to near perfection using the soft mask head during training. In addition, by using additional keypoints, our approach improves the localization detection and preserves its tightness among scales.
In contrast to the unchanged parking layout of the PKLot dataset, our SNU-SPS creates more challenges for the model to select the correct anchor patch due to its various capture positions. This makes not only our model but also other detectors struggle. Therefore, we omit the classifier method for a fair comparison. In this dataset, both the reference keypoints and the soft mask generator are not much efficient as the model grid is not stationary. Despite the challenge of adaptability, when looking into the heatmap of the soft mask through different scales of the spatial module in Fig \ref{fig: loc_heatmap}, the obtained mask on the parking lot still has denser attention than other foreground predictions. Combined with the impact of reference keypoints, OcpDet leaves a gap of nearly 5\% on mAP(0.5) to original MBN-FPN and 10\% to MBN. Moreover, in the first row of Fig \ref{fig: loc_heatmap}, it shows that our model is not sensitive with car appearance. The prediction only activates inside in the parking zone where the lines are visible.
\subsection{Result Filter Performance}
To evaluate the result filter efficiency, we solely address OcpDet on SNU-SPS as there is still room for model improvement. We calculate the error of an inference sample from the formula Eq \ref{eq: Filter} and remove a maximum of 100 samples (20\% of the test set) from the test set. For the purpose of testing out the benefit from the spatial module and the training error module, we lock the $\gamma$ by 0.7 and set $\alpha$ from a set of $[0,0.4,1]$ with respect to $OcpDet_{ll}$, $OcpDet_{both}$ and $OcpDet_{spatial}$ in Table \ref{tab: performance}. According to the experiment, the result filter has boosted the model's overall accuracy, proving that the filter can ensure a better quality from the detector regardless of $\alpha$ assignment. Because the spatial filter can score and remove unreliable results from detections by overlapping with the soft mask, its results are slightly better than the \cite{yoo2019learning} approach on the training filter.

\subsection{Optimal Routing and Parking Assignment}

To make a comprehensive benchmark on the impact of detection results on the assignment application layer, we collect the traffic information over ten days from the government website and associate that information with the test set to form a close loop simulation. Each day from 3 to 6pm, there will be about 100 requirements for booking a vacant spot to 6 parking lots. The suggested optimal road for each request will be assigned from the MapQuest API. We consider the Hungarian assignment \cite{kuhn1955hungarian} from the masked-out test label as the ground truth for assigning vacant spots. 

\begin{wrapfigure}[14]{r}{0.55\textwidth}
\includegraphics[width=\linewidth,trim={0 1.cm 1.5cm 2cm},clip]{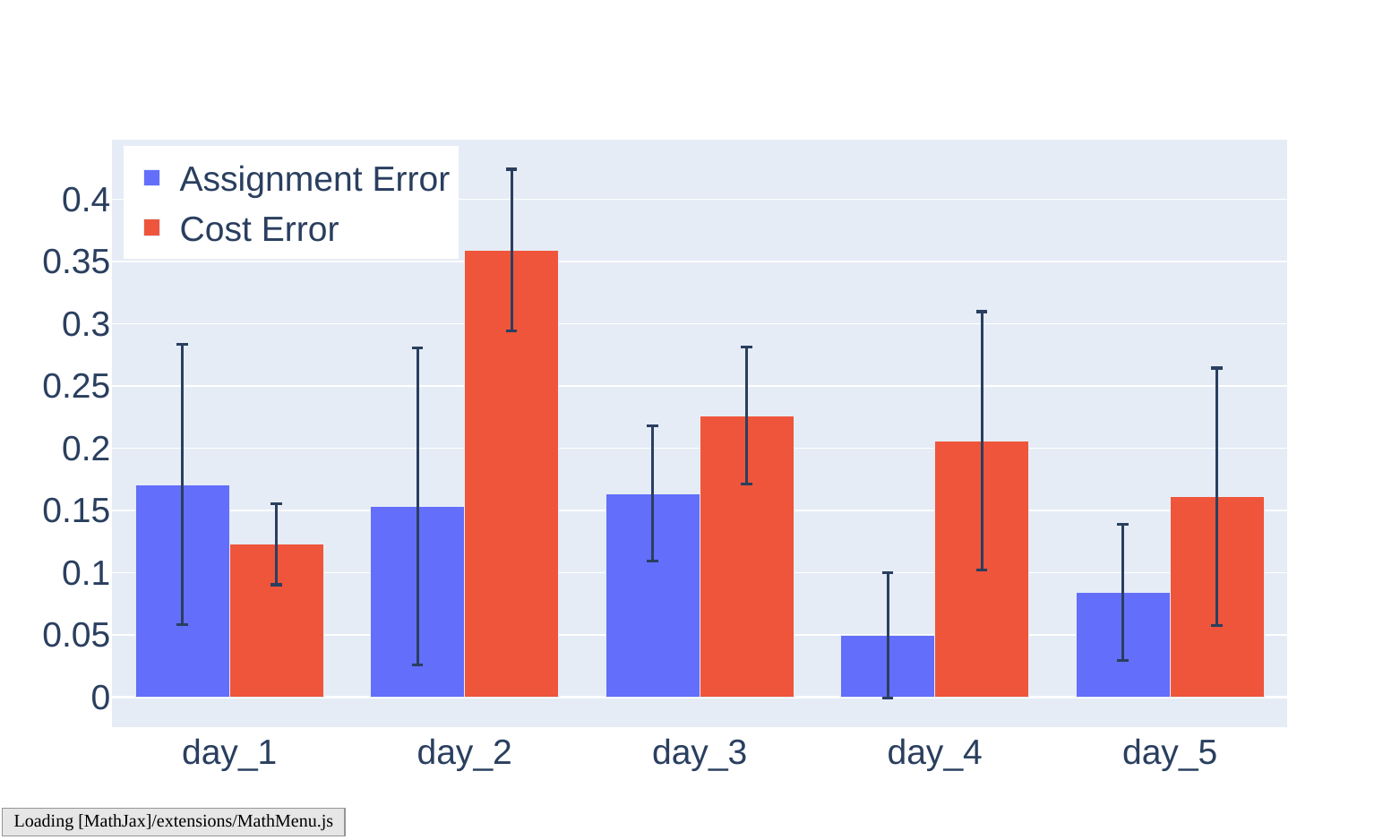}
\caption{The cost errors and the assignment errors: averaging simulation for 5 days at 6 parking lots}
\label{fig: assign}
\end{wrapfigure}
% As mentioned in section \ref{sect: parking_assignment}, the assignment only focuses on the vacant spot in a parking lot. Therefore, only masked-out vacant predictions are serialized to the parking assignment task. 
Then, we compute two evaluation metrics: cost error and assignment error. The cost error is computed by the absolute error of the ground truth assignment cost $C_{g,i}$ and the vacancy detection assignment cost $C_{p,i}$, which is designed by Eq. \ref{eq: Cost}. The assignment cost is computed after getting the assignments. From each assigned parking lot in 6 parking lots, we compute the total number of booked slots $N$ and compare those between the ground truth $ N_{g,i,j}$ and the detection $ N_{p,i,j}$ by an absolute error. We perform normalization to keep the value from 0 to 1. The simulation will repeat ten times to capture the model's average performance due to different traffic statuses. 
\begin{gather}
    \label{eq: Cost}
    C = \gamma C_{price} + (1-\gamma)(C_{travel} + C_{distance})\\
    Err_{cost} = \sum_{i=3}^{6} \frac{\mid C_{g,i} - C_{p,i}\mid}{C_{g,i}} \quad \quad
    Err_{assign} = \sum_{i=3}^{6} \sum_{j=1}^{6} \frac{\mid N_{g,i,j} - N_{g,i,j} \mid}{N_{g,i}}
\end{gather}
From Fig. \ref{fig: assign}, OcpDet allows the system to operate at most 40\% error for the cost-minimizing budget while maintaining at least 70\% correct on assignment. Because the cost error is related to the distance travel, wrong assignments placed on far-distance drives can cause a huge gap to the optimal cost. But this error is not much in terms of matching the number of assignments. From these metrics, we can overview the system's benefits to users. Operators will benefit the most as their vacant spaces will automatically be assigned with minima error. In contrast, some drivers may get some disadvantages from the system assignment.
\section{Conclusion \& Future work}
This paper proposes a novel end-to-end CV-SPS with a detailed benchmark on both old and new datasets. Even though our dataset is small, it shows challenging factors, and it is the first dataset for computer vision with full CV-SPS scope. Our method has proved its efficiency to some extent and can potentially close the gap to the classifier approach when addressing stationary views. Moreover, we also provide a novel filtering method that can help the system approach an error-free execution. In the future, we will continue building our dataset to a bigger scale with which it can provide functional information for SPS, such as vehicle reidentification or parking type selection.
\section{Acknowledgement}
Dataset and experiment in this work were supported by the Automotive Industry Building Program (1415177436, Building an open platform ecosystem for future technology innovation in the automotive industry) funded by the Ministry of Trade, Industry Energy (MOTIE, Korea).

% traditional approaches by using sensors tend to be much costly and limited.

\clearpage
\bibliographystyle{splncs04}
\bibliography{parking}
\end{document}